\documentclass[conference]{IEEEtran}
\IEEEoverridecommandlockouts
\usepackage{cite}
\usepackage{amsmath,amssymb,amsfonts}
\usepackage{mathrsfs}
\usepackage{algorithm}
\usepackage{algpseudocode}
\usepackage{graphicx}
\usepackage{textcomp}
\usepackage{xcolor}
\usepackage[inkscapelatex=false]{svg}

\def\BibTeX{{\rm B\kern-.05em{\sc i\kern-.025em b}\kern-.08em
    T\kern-.1667em\lower.7ex\hbox{E}\kern-.125emX}}
\begin{document}

\title{Compact and Efficient Neural Networks for Image Recognition Based on Learned 2D Separable Transform\\
\thanks{The authors express their sincere gratitude to the resident of the High-Tech Park, the “Engineering Center Yadro” company, which is one of the YADRO development centers, for providing equipment for conducting experiments within the framework of the joint educational laboratory with the Belarusian State University of Informatics and Radioelectronics.}
}


\author{\IEEEauthorblockN{Maxim Vashkevich\,\IEEEauthorrefmark{2} 
and Egor Krivalcevich\,\IEEEauthorrefmark{1}}
\IEEEauthorblockA{Department of Computer Engineering\\ Belarusian State University of Informatics and Radioelectronics\\
6, P.~Brovky st., 220013, Minsk, Belarus\\
Email: \IEEEauthorrefmark{2}vashkevich@bsuir.by, \IEEEauthorrefmark{1}krivalcevi4.egor@gmail.com}}

\maketitle

\begin{abstract}
The paper presents a learned two-dimensional separable transform (LST) that can be considered as a new type of computational layer for constructing neural network (NN) architecture for image recognition tasks. The LST based on the idea of sharing the weights of one fullyconnected (FC) layer to process all rows of an image. After that, a second shared FC layer is used to process all columns of image representation obtained from the first layer. 
The use of LST layers in a NN architecture significantly reduces the number of model parameters compared to models that use stacked FC layers.
We show that a NN-classifier based on a single LST layer followed by an FC layer achieves 98.02\% accuracy on the MNIST dataset, while having only 9.5k parameters. We also implemented a LST-based classifier for handwritten digit recognition on the FPGA platform to demonstrate the efficiency of the suggested approach for designing a compact and high-performance implementation of NN models.
\end{abstract}


\section{Introduction}
Compact and high-performance implementations of neural networks (NN) are of great interest~\cite{Park-2016,Huynh-17}. Field Programmable Gate Arrays (FPGAs) usually considered as a energy efficient and flexible platform for NN implementation~\cite{Liang-2018}.
However, one of the main problem of FPGA-based NN implementation is a lack of memory resources on this type of computational platforms~\cite{Park-2016,Liang-2018}. It is known that NNs  have a large number of parameters that must be stored in memory. Many of the designs proposed in the literature used external DDR/SRAM memory to store the weights of the models~\cite{Park-2016,Liang-2018,Han-2020,Nakahara-2016}. 

There are several ways to reduce the number of parameters and memory resources usage: 1)\;quantization; 2)\;pruning (or increasing the sparseness of weights); 3)\;weights sharing.
Quantization effectively optimizes the memory usage by reducing the bit-representation of NN weights. The pruning process sets a portion of the NN's weights to zero, therefore, there is no need to store them in memory. Weights-sharing is a technique that involves reusing the weights of NNs during the inference process, thereby reducing the number of model parameters. For example, fullyconnected (FC) layers contribute to most of the weights~\cite{Qiu-2016}, but used only once in one inference process. On the contrary, the weights of the convolutional layers contribute less to the total number of model parameters, but reused in the inference process.

In this paper, we introduce a learned two-dimensional separable transform ($\mathrm{LST_{2D}}$) that can be considered as a new type of NN layer that follows the weights-sharing concept. $\mathrm{LST_{2D}}$ based on the idea of sharing the weights of one FC layer to process all rows of an image. After that, a second shared FC layer is used to process all columns of image representation obtained after the first layer.

We investigated LST in the context of the MNIST handwritten digit classification task. We showed that LST can be used to construct deep neural network (DNN) architectures, replacing the traditional feedforward neural network (FFNN). The NN based on LST allows one to achieve high accuracy while the number of model parameters is considerably lower than in FFNN.

The rest of this paper is organized as follows: section II provides brief overview on FFNN and convolutional NN, which are widely used in image recognition tasks. Section III describes the proposed learned two-dimensional transform and several NN architectures based on this transform. Section IV presents FPGA implementation of one block LST model for handwritten digit classification. Section V describes the experimental evaluation, and Section VI concludes the paper.
\begin{figure*}[!h]
\centering
\includesvg[width=0.99\linewidth]{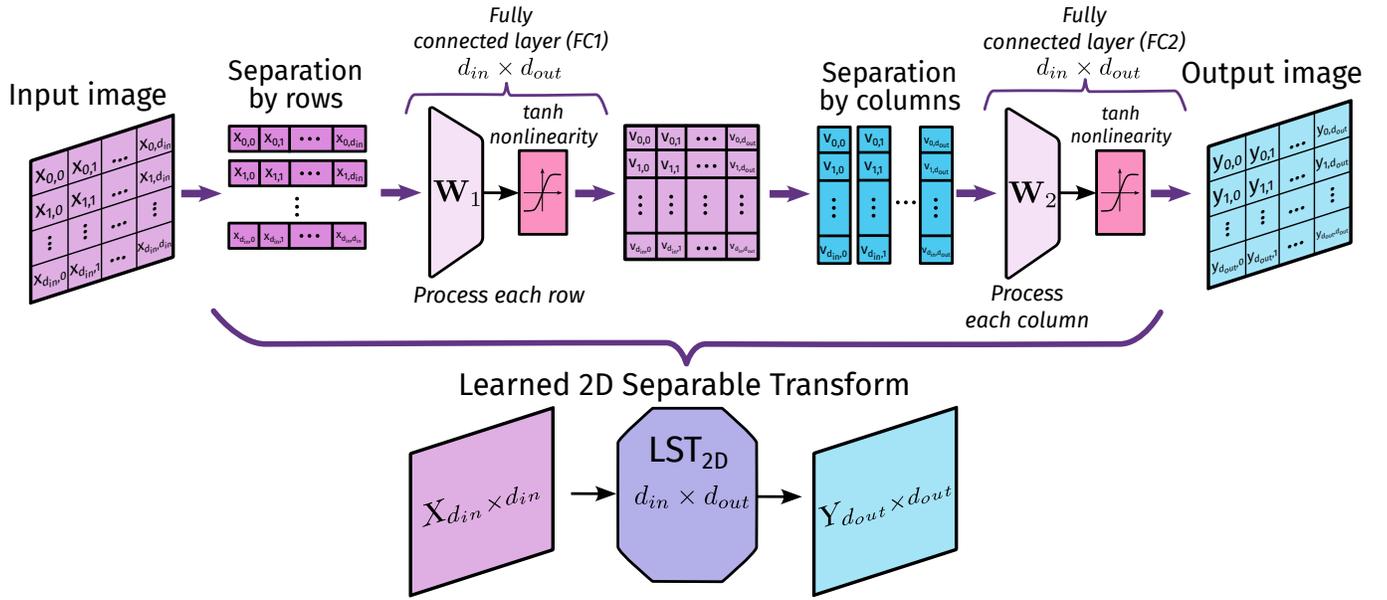}
\caption{Structure of basic block of learned separable 2D transform}
\label{fig:SLT_one_block}
\end{figure*}

\section{NN for handwritten digit recognition}
In this section, we will discuss the two most common NN architectures used for hardware implementations of image recognition tasks: FFNN and CNN.

\subsection{Fullyconnected NN}
FFNN has multiple hidden layers, each layer $k$ processes input vector $\mathbf{x}_k \in \mathbb{R}^{d_{in}^k}$, which propagated to the next layer using transform:

\begin{equation}
\mathbf{x}_{k+1} = \mathrm{FC}(\mathbf{x}_{k}) = g(\mathbf{W}_k \mathbf{x}_{k}),
\label{eq:FC}
\end{equation}
where $\mathrm{FC}(\cdot)$ denotes the function of FC layer, $\mathbf{W}_k \in \mathbb{R}^{d_{out}^k} \times \mathbb{R}^{d_{in}^k}$ -- weight matrix, $g(\cdot)$ is non-linear activation function. Note, that in~\eqref{eq:FC} bias terms  are incorporated into weight matrix.

Usually in FFNN architectures used for handwritten digit recognition, the input image of size $d_{in} \times d_{in}$ is reshaped to a one-dimensional vector of length $d_{in}^2$. This transformation is called \emph{flattening}.

The pseudocode for the computation of the FFNN for the recognition of handwritten digits is given in the algorithm~\ref{alg:DNN}.

\begin{algorithm}
\caption{FFNN computation}\label{alg:DNN} 
\begin{algorithmic}
\State $\mathbf{x}_1 \gets \mathbf{flatten}(\mathbf{X}_{d_{in}\times d_{in}})$
\For{$k$ in $1$ to $N-1$}
    \State $\mathbf{x}_{k+1} \gets g(\mathbf{W}_k \mathbf{x}_{k})$
\EndFor
\State $\mathbf{y} \gets \mathrm{softmax}(\mathbf{W}_N \mathbf{x}_{N-1})$ 
\end{algorithmic}
\end{algorithm}

FFNN architecture is widely used for FPGA implementation~\cite{Park-2016, Huynh-17, Liang-2018, Medus-2019} due to its simplicity and structural regularity.

\subsection{Convolutional NN}

Convolutional Neural Networks (CNNs) are a type of deep neural network primarily used for processing image data. A CNN consists of multiple convolutional layers that apply a series of learnable filters to the input image.

In a multilayer CNN, each layer receives input from a small region of the previous layer's output. This approach significantly reduces the number of model parameters and enables the detection of local features (such as corners, edges, and others). Essentially, a 2D convolutional layer can be viewed as a matrix operation, followed by an activation function. Typical CNN architecture is built from common components: convolutional layers, pooling layers, and FC layers (applied at the output).

The key feature of CNNs is the use of the weight sharing concept in convolutional layers, which reduces the number of learnable parameters. This allows for hierarchical learning of features, from low-level to high-level representations. However, CNNs are more complex from the implementation point of view, compared to FFNNs.

\section{NN architecture based on learned 2D separable transform ($\mathrm{LST_{2D}}$)}

The suggested learned 2D separable transform ($\mathrm{LST_{2D}}$) processes the image in a row-by-column manner instead of converting it into a vector, as is usually done in FFNN. The pseudocode for $\mathrm{LST_{2D}}$ is given below.

\begin{algorithm}
\caption{$\mathrm{LST_{2D}}$: building block}\label{alg:L2DST} 
\begin{algorithmic}
\For{$k$ in $1$ to $d_{in}$}
    \State $\mathbf{V}[k,:] \gets \mathrm{FC1}(\mathbf{X}[k,:])$
\EndFor
\For{$k$ in $1$ to $d_{in}$}
    \State $\mathbf{Y}[:,k] \gets \mathrm{FC2}(\mathbf{V}[:,k])$
\EndFor
\end{algorithmic}
\end{algorithm}

From a mathematical point of view, $\mathrm{LST_{2D}}$ takes the 2D image $\mathbf{X}$ of size $d_{in}\times d_{in}$ as input and produces the 2D output image $\mathbf{Y}$ of size $d_{out}\times d_{out}$: 

\begin{equation}
\mathbf{Y} = \mathrm{LST}_{d_{in}\times d_{out}}(\mathbf{X}) = \tanh(\mathbf{W}_2 \tanh(\mathbf{W}_1 \mathbf{X}^T)),
\label{eq:L2DST}
\end{equation}
where $\mathbf{W}_1$, $\mathbf{W}_2$ are the weight matrices of the $\mathrm{FC1}$ and  $\mathrm{FC2}$ layers, respectively (see the algorithm~\ref{alg:L2DST} and Fig.~\ref{fig:SLT_one_block}).
Here $d_{in}$ and $d_{out}$ are the hyper-parameters of the transform, that determines the number of learnable parameters, that equal to $N = 2\cdot(d_{in} + 1)\cdot d_{out}$. We use notation $\mathrm{LST}_{d_{in}\times d_{out}}$ to specify the layer of learned separable transform that takes image of size $d_{in}\times d_{in}$ and produces the $d_{out}\times d_{out}$ image. Also we use terms $\mathrm{LST_{2D}}$ or LST when speak about proposed transform in general.

\begin{figure}
\centering
\includesvg[width=0.85\linewidth]{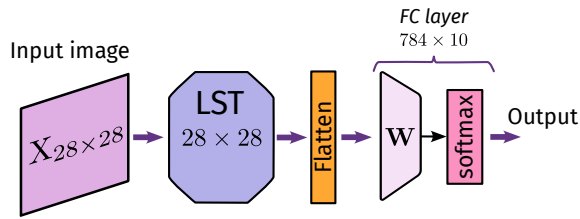}
\caption{One-block LST NN architecture (model $\mathrm{LST_{2D}}$-1)}
\label{fig:One-block_L2DST_NN}
\end{figure}

The simplest NN architecture based on $\mathrm{LST_{2D}}$ for MNIST digit recognition requires one $\mathrm{LST_{2D}}$ block followed by an FC layer with a softmax activation function. This architecture is presented in Fig.~\ref{fig:One-block_L2DST_NN}, we refer to this model as LST-1 (or $\mathrm{LST_{2D}}$-1).

\begin{figure}
\centering
\includesvg[width=\linewidth]{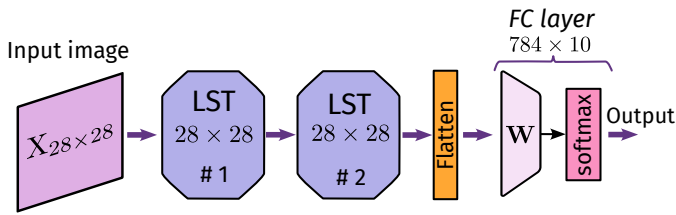}
\caption{Two-block LST NN architecture (model $\mathrm{LST_{2D}}$-2)}
\label{fig:Two-block_L2DST_NN}
\end{figure}

We can stack $\mathrm{LST_{2D}}$ blocks to create deep neural network architectures that have greater expressive power and grater number of learnable parameters. An example of two-block $\mathrm{LST_{2D}}$ NN architecture is given in Fig.~\ref{fig:Two-block_L2DST_NN}. We refer to the model given in Fig.~\ref{fig:Two-block_L2DST_NN} as LST-2 (or $\mathrm{LST_{2D}}$-2).

Model LST-2 has an additional parameter $d_h$, which defines the dimensionality $d_h\times d_h$ of the hidden representation of the input image.

\begin{figure}
\centering
\includesvg[width=0.6\linewidth]{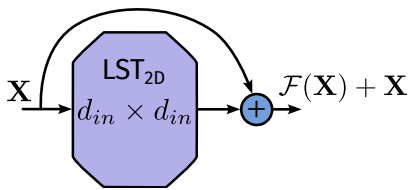}
\caption{ResNet-like L2DST: building block}
\label{fig:Res-L2DST}
\end{figure}

The suggested $\mathrm{LST_{2D}}$ blocks can also be used to construct ResNet-like structures with skip connections~\cite{He-2016}. To do this, the dimensionality of the hidden representation should be set equal to the input dimensionality, that is, $d_h=d_{in}$. An example $\mathrm{LST_{2D}}$ block with skip connection is given in Fig.~\ref{fig:Res-L2DST}. Therefore, LST can be used as a building block for creating a deep neural network architecture.

\section{FPGA implementation}
\begin{figure*}
\centering
\includesvg[width=0.80\linewidth]{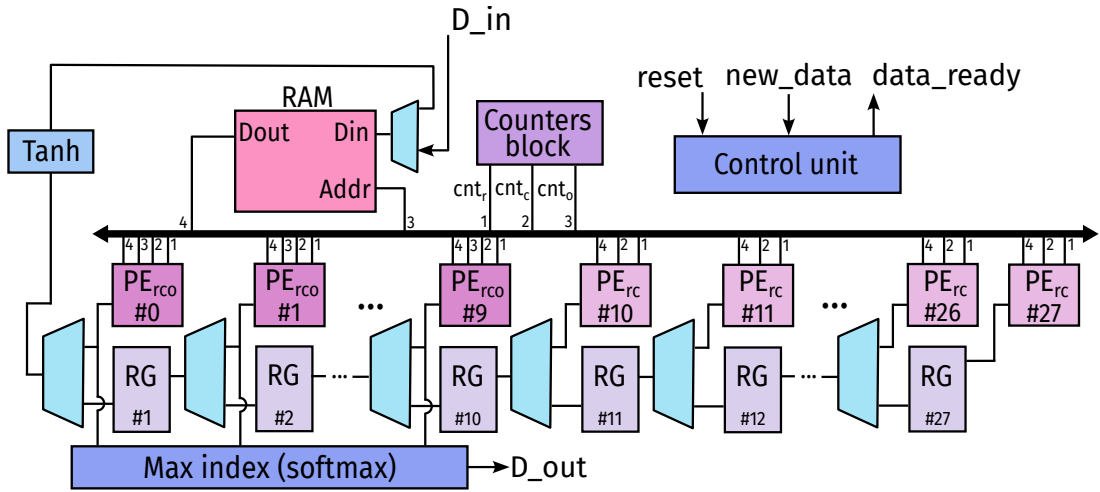}
\caption{Hardware implementation of the LST-1 model}
\label{fig:FPGA_system_overall}
\end{figure*}

In this section we consider the hardware architecture of FPGA implementation of the LST-1 model (see Fig.~\ref{fig:One-block_L2DST_NN}). The proposed architecture has a single physical computing layer to perform all needed computation. The detailed architecture of LST-1 model is shown in Fig.~\ref{fig:FPGA_system_overall}. 

\begin{figure}
\centering
\includesvg[width=0.55\linewidth]{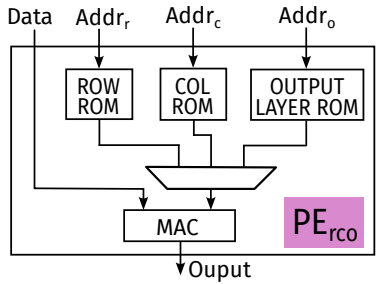}
\caption{Processing element architecture (row/column/output)}
\label{fig:PE_rco}
\end{figure}

\begin{figure}
\centering
\includesvg[width=0.40\linewidth]{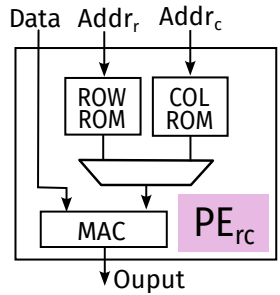}
\caption{Processing element architecture (row/column)}
\label{fig:PE_rc}
\end{figure}

The architecture consists of two types of processing elements (PE). PE$_{rco}$ (Fig.~\ref{fig:PE_rco}) is a processing element that is used to calculate the FC layers for the rows, columns and the output FC layer. PE$_{rc}$ (Fig.~\ref{fig:PE_rc}), on the other hand, supports the calculation of FC layers for rows and columns. Each PE consists of multiplication and accumulation (MAC) core and ROM units that stores one row of weight matrix of particular FC layers.

At the first stage, the input image is sent to the \texttt{D\_in} input and stored in RAM memory. At the second stage, the image rows are sequentially read from RAM and processed on PEs using FC1 weights stored in ROM \texttt{ROW}. The results from PEs are passed to registers \texttt{RG} and sequentially passed through the tanh block and
stored in RAM. Therefore, the suggested architecture adopts the concept of in-place computation. Also, this approach allows to implement only one block for activation function calculation that leads to savings of hardware resources. After processing all rows of the image, the RAM memory contains the output of the FC1 layer (see Fig.~\ref{fig:SLT_one_block}). At the third stage, the content of the RAM processed on PEs by columns using FC2 weights stored in ROM \texttt{COL}. After processing all columns, the RAM memory contains the output of the FC2 layer. In the fourth stage of the computation, the image representation obtained after LST is read from the RAM and processed using 10 PEs (of $rco$-type) that have ROMs with weights of the FC output layer. At the final stage, the index of the maximum element is searched among the outputs of the 10 PEs. This index indicates the digit that the model has recognized. Final stage implemented in \texttt{Max index} block (see Fig.~\ref{fig:FPGA_system_overall}).

\texttt{Tanh} block implements calculation of hyperbolic tangent function.
For our design, we used the following approximation of $\mathrm{tanh}$ proposed in~\cite{Medus-2019}:

\begin{equation}
\mathrm{tanh}(x)\approx F(x) = 
  \begin{cases}
  \mathrm{sign}(x),         & \mathrm{if}\;\;  |x|>2, \\
  (1 + \frac{x}{4})\cdot x, & \mathrm{if}\;\; -2<x<0, \\
  (1 - \frac{x}{4})\cdot x, & \mathrm{if}\;\;  0<x<2. \\
  \end{cases}
\end{equation}

\section{Experimental results}
\subsection{Dataset}
The MNIST dataset is a benchmark dataset widely used for handwritten digit recognition tasks. It consists of grayscale images of digits ranging from 0 to 9, with each image normalized and centered in a 28x28 pixel grid. The dataset is divided into two subsets: a training set containing 60\;000 samples and a test set with 10\;000 samples. Each image in the dataset represents a single handwritten digit, with pixel intensity values ranging from 0 (black) to 255 (white). We used all 784 image pixels as input neurons to the network. The LST-based models processes these images by first taking the 28×28 image as input and producing an output of the same size, 28×28, after applying a series of transformations. The output matrix is then flattened into a 1D vector of size 1×784. This flattened vector is passes through a FC layer followed by a softmax activation function.

The softmax layer generates a probability distribution across 10 classes (digits 0 to 9). The class with the highest probability indicates the digit present in the input image.

The training of the models has been performed using the PyTorch library. 
All models are trained for 300 epochs. The initialization of the weights is performed using Glorot method~\cite{Glorot-2010}.
The batch size is fixed to 1000 and training is performed using the Adam optimizer~\cite{Kingma2014} with with the learning rate equal to 2e-3 and weight decay 1e-5.

\subsection{Experimental results}
The first experiment that was conducted has the objective of estimating the performance of the classifier based on one-block $\mathrm{LST_{2D}}$ NN (see Fig.~\ref{fig:One-block_L2DST_NN}), we refer to this model as $\mathrm{LST_{2D}}$-1. The accuracy achieved is 98.02\%, which is very close to 98.16\% reported in ~\cite{Huynh-17}. However, the number of parameters in the $\mathrm{LST_{2D}}$-1 model is 12.2 times smaller than in the DNN proposed in~\cite{Huynh-17}.

\begin{figure}[!h]
\centering
\includesvg[width=\linewidth]{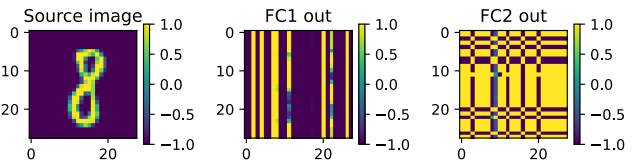}
\caption{Representation of the handwritten digit at the hidden layers of the $\mathrm{LST_{2D}}$-1 model}
\label{fig:L2DST_embedding}
\end{figure}

Fig.~\ref{fig:L2DST_embedding} shows the representation of handwritten digit eight obtained at the hidden layers of the $\mathrm{LST_{2D}}$-1 model.

It can be seen that the $\mathrm{LST_{2D}}$-1 model encoded the image as an irregular chessboard-like pattern.

The second experiment that was conducted has the objective of estimating the performance of the classifier based on two-block $\mathrm{LST_{2D}}$ NN (see Fig.~\ref{fig:Two-block_L2DST_NN}). The accuracy achieved is 98.34\%, which is slightly higher than 98.32\% reported in~\cite{Liang-2018}. However, the number of parameters in the $\mathrm{LST_{2D}}$-2 model is 901.9 times smaller than in the DNN proposed in~\cite{Liang-2018}.

\begin{figure}[!h]
\centering
\includesvg[width=\linewidth]{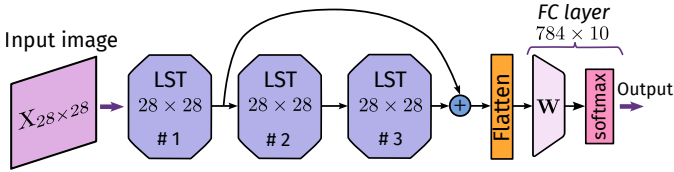}
\caption{Model ResLST-3 (with skip connection)}
\label{fig:ResLST-3}
\end{figure}

In the third experiment, we constructed ResNet-like NN architecture based on three LST units (see Fig.~\ref{fig:ResLST-3}). We call this model ResLST-3. 

The accuracy of the ResLST-3 model is 98.53\%, which is higher than 98.4\% achieved by the 3-layer DNN reported in~\cite{Umuroglu-2017}.
Again, it is worth mentioning that NN based on LST units has considerably less parameters than feedforward DNN based on FC layers. For example, the ResLST-3 model has 146 times less parameters than the 3-layer DNN (LFC-max)~\cite{Umuroglu-2017}.

Table~\ref{tab:compare} summarizes three NN architectures based on LST units suggested in this work with several known DNN architectures for handwritten digit recognition. 
\begin{table}[!h]
\setlength\tabcolsep{3pt}
\caption{Comparison of the proposed $\mathrm{LST_{2D}}$-based NN architectures with other DNN}
\begin{center}
\begin{tabular}{|c|c|c|c|c|c|}
\hline
Authors \& ref. & DNN architecture & \# Params & Accuracy \\
\hline
Huynh~\cite{Huynh-17} & 784-40-40-40-10 & 34\;960 & 97.20\% \\
Huynh~\cite{Huynh-17} & 784-126-126-10 & 115\;920 & 98.16\% \\
Westby, et al.~\cite{Westby-2021} & 784-12-10 & 9\;550 & 93.25\% \\
Umuroglu, et al.~\cite{Umuroglu-2017} & 784-1024-1024-10 & 1\;863\;690 & 98.40\% \\
Medus, et al.~\cite{Medus-2019} & 784-600-600-10 & 891\;610 & 98.63\% \\
Liang, et al.~\cite{Liang-2018} & 784-2048-2048-2048-10 & 10\;100\;000 & 98.32\% \\
$\mathrm{LST_{2D}}$-1 [proposed] & $\mathrm{LST_{28\times28}}$-784-10 & 9\;474 & 98.02\% \\
$\mathrm{LST_{2D}}$-2 [proposed] & see Fig.~\ref{fig:Two-block_L2DST_NN} & 11\;098 & 98.34\% \\
ResLST-3 [proposed] & see Fig.~\ref{fig:ResLST-3} & 12\;722 & 98.53\% \\
\hline
\end{tabular}
\label{tab:compare}
\end{center}
\end{table}

From the table~\ref{tab:compare} it can be seen that to achieve  the accuracy above 98\%, a DNN with at least 100K parameters is required. 
At the same time, the suggested $\mathrm{LST_{2D}}$-1 model achieves such accuracy, while having only 9.5K parameters. The DNN proposed in~\cite{Westby-2021} has almost the same number of parameters as the $\mathrm{LST_{2D}}$-1 model, however, its accuracy is considerably lower.


\subsection{FPGA implementation}
The hardware realization of LST-1 model is designed using Xilinx Vivado (v.2023.2), and implemented on a Xilinx Zybo Z7 evaluation board. This board contains XC7Z010 which has ARM CPU and programmable logic. To simplify development and testing process on this platform, the Linux PINQ (Python productivitY for zyNQ)  distribution is used. PINQ allows to interact with FPGA hardware units implemented as IP cores using a Jupyter notebook software, which makes the testing and development process more flexible and convenient. So, we have implemented the LST-1 model as an IP core. We used a 12-bit representation for the weights and internal data. Five bits were used to represent the integer part and seven bits for the fractional part. The implementation of proposed LST-1 model requires 6473 LUTs (36.8\%), 680 Flip-Flops (1.9\%) and 29 RAMB18 (24.2\%). We did not estimate the throughput or other performance metrics for the suggested design. Our primary goal was to evaluate the effects of weight quantization on model accuracy. We found that the accuracy of the model when implemented on an FPGA remained the same as for a floating-point model. Therefore, we conclude that the proposed approach to constructing NN based on LST can be used to design new hardware architectures for image classification tasks.

\section{Discussion and future work}
The proposed LST can be considered as building block for constructing NN for image recognition. Particularly we showed that NN based on LST is a good alternative to FFNN, that traditionally used for designing FPGA implementation for handwritten digits recognition~\cite{Park-2016, Huynh-17, Medus-2019}. Application of LST units significantly reduces memory consumption while maintaining high performance. The proposed approach of designing NN based on LST can be also applied to other gray-scale image recognition tasks (such as Fashion MNIST and etc.). However, the investigation of the performance of the proposed approach in context of other gray-scale image recognition tasks is the subject of future works. The extension of proposed approach to processing and recognition of the colour image is a challenging task, because images are inherently 3D data, while the LST operates on 2D input. One of the possible solution to this problem is to use quaternionic neural network. With this approach we can represent the colour image as a 2D matrix of quaternions that can be processed by quaternion-valued fullyconnected layers~\cite{Petrovsky-2024}.

\section{Conclusion}
We suggested learnable two-dimensional separable transform (LST) that is based on the weight-sharing concept of two FC layers -- one for processing rows and second for processing columns. The designed LST-1 and LST-2 models, that used, correspondingly one and two LST units, allows to achieve high accuracy (98.02\% and 98.34\%) while have more than 10 times less number of parameters than DNNs with similar accuracy. Fixed-point implementation of the LST-1 model on FPGA shows that quantization of the model weights not leads to drop of accuracy. Therefore, the proposed approach to constructing NN based on LST can be used to design compact and efficient hardware architectures for image classification tasks.



\bibliographystyle{IEEEtran}

\begin{thebibliography}{10}
    
    \bibitem{Park-2016}
    J.~Park and W.~Sung, ``{FPGA} based implementation of deep neural networks
      using on-chip memory only,'' in \emph{2016 IEEE International conference on
      acoustics, speech and signal processing (ICASSP)}, 2016, pp. 1011--1015.
    
    \bibitem{Huynh-17}
    T.~V. Huynh, ``Deep neural network accelerator based on {FPGA},'' in \emph{2017
      4th NAFOSTED Conference on Information and Computer Science}, 2017, pp.
      254--257.
    
    \bibitem{Liang-2018}
    S.~Liang, S.~Yin, L.~Liu, W.~Luk, and S.~Wei, ``{FP-BNN}: Binarized neural
      network on {FPGA},'' \emph{Neurocomputing}, vol. 275, pp. 1072--1086, 2018.
    
    \bibitem{Han-2020}
    J.~Han, Z.~Li, W.~Zheng, and Y.~Zhang, ``Hardware implementation of spiking
      neural networks on {FPGA},'' \emph{Tsinghua Science and Technology}, vol.~25,
      no.~4, pp. 479--486, 2020.
    
    \bibitem{Nakahara-2016}
    H.~Nakahara, H.~Yonekawa, T.~Sasao, H.~Iwamoto, and M.~Motomura, ``A
      memory-based realization of a binarized deep convolutional neural network,''
      in \emph{2016 International Conference on Field-Programmable Technology
      (FPT)}, 2016, pp. 277--280.
    
    \bibitem{Qiu-2016}
    J.~Qiu, J.~Wang, S.~Yao, K.~Guo, B.~Li, E.~Zhou, J.~Yu, T.~Tang, N.~Xu, S.~Song
      \emph{et~al.}, ``Going deeper with embedded {FPGA} platform for convolutional
      neural network,'' in \emph{Proceedings of the 2016 ACM/SIGDA international
      symposium on field-programmable gate arrays}, 2016, pp. 26--35.
    
    \bibitem{Medus-2019}
    L.~D. Medus, T.~Iakymchuk, J.~V. Frances-Villora, M.~Bataller-Mompe{\'a}n, and
      A.~Rosado-Munoz, ``A novel systolic parallel hardware architecture for the
      {FPGA} acceleration of feedforward neural networks,'' \emph{IEEE Access},
      vol.~7, pp. 76\,084--76\,103, 2019.
    
    \bibitem{He-2016}
    K.~He, X.~Zhang, S.~Ren, and J.~Sun, ``Deep residual learning for image
      recognition,'' in \emph{Proceedings of the IEEE conference on computer vision
      and pattern recognition}, 2016, pp. 770--778.
    
    \bibitem{Glorot-2010}
    X.~Glorot and Y.~Bengio, ``Understanding the difficulty of training deep
      feedforward neural networks,'' in \emph{Proceedings of the thirteenth
      international conference on artificial intelligence and statistics}.\hskip
      1em plus 0.5em minus 0.4em\relax JMLR Workshop and Conference Proceedings,
      2010, pp. 249--256.
    
    \bibitem{Kingma2014}
    D.~P. Kingma and J.~Ba, ``Adam: A method for stochastic optimization,''
      \emph{arXiv preprint arXiv:1412.6980}, 2014.
    
    \bibitem{Umuroglu-2017}
    Y.~Umuroglu, N.~J. Fraser, G.~Gambardella, M.~Blott, P.~Leong, M.~Jahre, and
      K.~Vissers, ``{FINN}: A framework for fast, scalable binarized neural network
      inference,'' in \emph{Proceedings of the 2017 ACM/SIGDA international
      symposium on field-programmable gate arrays}, 2017, pp. 65--74.
    
    \bibitem{Westby-2021}
    I.~Westby, X.~Yang, T.~Liu, and H.~Xu, ``{FPGA} acceleration on a multi-layer
      perceptron neural network for digit recognition,'' \emph{The Journal of
      Supercomputing}, vol.~77, no.~12, pp. 14\,356--14\,373, 2021.
    
    \bibitem{Petrovsky-2024}
    N.~Petrovsky and M.~Vashkevich, ``Quaternionic multilayer bottleneck
      autoencoder for color image compression,'' in \emph{2024 Signal Processing:
      Algorithms, Architectures, Arrangements, and Applications (SPA)}, 2024, pp.
      19--23.
    
    \end{thebibliography}

\vspace{12pt}

\end{document}